\pdfoutput=1

\documentclass[11pt]{article}

\usepackage[]{acl}

\usepackage{times}
\usepackage{latexsym}

\usepackage{latexsym}
\usepackage{epsfig}
\usepackage{graphicx}
\usepackage[linesnumbered,ruled,vlined]{algorithm2e}
\usepackage{xcolor}

\usepackage{multirow}
\usepackage{etoolbox}
\usepackage{balance}
\usepackage{amsmath}
\usepackage{amsfonts}
\usepackage{balance}
\usepackage{pifont}
\usepackage{makecell}
\usepackage{hhline}
\usepackage{booktabs}
\usepackage{booktabs}
\usepackage{makecell}
\usepackage{graphicx}
\usepackage{mathtools}
\usepackage{enumitem}
\usepackage{float}
\usepackage{colortbl}
\usepackage[most]{tcolorbox}
\usepackage{pifont}
\usepackage{subcaption}
\usepackage{hyperref}

\usepackage[T1]{fontenc}

\usepackage[utf8]{inputenc}

\usepackage{microtype}

\title{Chaining Event Spans for Temporal Relation Grounding}

\newcommand\correspondingauthor{\thanks{~~Corresponding author.}}
\author{Jongho Kim$^{\spadesuit}$$^{\clubsuit}$, Dohyeon Lee$^{\spadesuit}$, Minsoo Kim$^{\spadesuit}$$^{\clubsuit}$, Seung-won Hwang$^{\spadesuit}$$^{\clubsuit}$\correspondingauthor \\
$^{\spadesuit}$Seoul National University \\
$^{\clubsuit}$ Interdisciplinary Program in Artificial Intelligence, Seoul National University \\ 
\texttt{\{jongh97, waylight3, minsoo9574, seungwonh\}@snu.ac.kr} \\
}

\newcommand{\ours}{TRN\xspace}
\newcommand{\batchattn}{Evidence Chaining Step\xspace}
\newcommand{\graph}{Evidence Extraction Step\xspace}

\newcommand{\jonghoo}[1]{{\color{black}#1}}

\begin{document}
\maketitle
\begin{abstract}

Accurately understanding temporal relations between events
is a critical building block of diverse tasks, such as temporal reading comprehension (TRC) and relation extraction (TRE).
For example in TRC, we need to understand the temporal semantic differences
between the following two questions that are lexically near-identical:
``What finished right before the decision?'' or ``What finished right after the decision?''.
To discern the two questions, existing solutions have relied on 
answer overlaps as a proxy label
to contrast similar and dissimilar questions.
However, we claim that answer overlap can lead to unreliable results, due to spurious overlaps of two dissimilar questions with coincidentally identical answers.
To address the issue, we propose a novel approach that elicits proper reasoning behaviors through a module for predicting time spans of events. 
We introduce the Timeline Reasoning Network (\ours) operating in a two-step inductive reasoning process: In the first step model initially answers each question with semantic and syntactic information.
The next step chains 
multiple questions on the same event to predict a timeline, which is then used to ground the answers.
Results on the TORQUE and TB-dense, TRC and TRE tasks respectively, demonstrate that \ours 
outperforms previous methods by effectively resolving the spurious overlaps
using the predicted timeline~\footnote{Codes are available here:~\url{https://github.com/JonghoKimSNU/Temporal-Reasoning-Network/tree/main}}.

\end{abstract}


\section{Introduction}
\label{intro}

\begin{figure*}[t]
    \centering
    \includegraphics[width=0.95\linewidth]{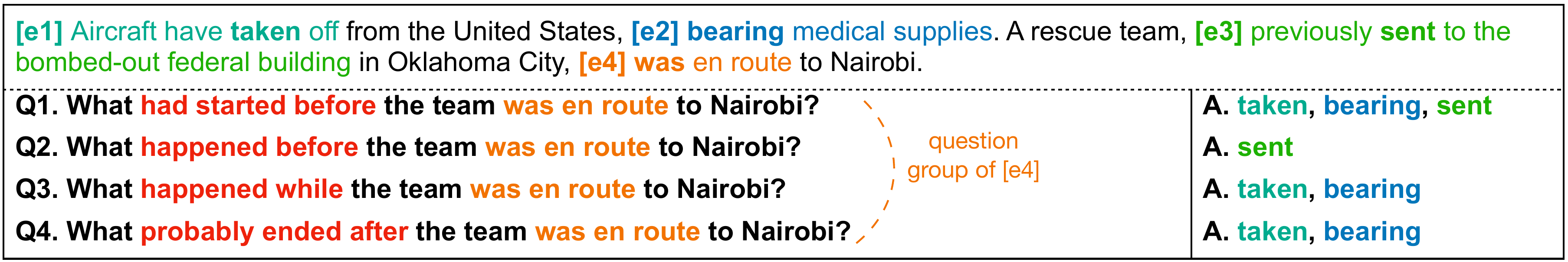}
    \vspace{-3mm}
    \caption{Example of passage and question grouped by the same event (`the team was en route'') in temporal reading comprehension. Events are highlighted in color and temporal relations in the questions are in red.}
    \vspace{-3mm}
    \label{fig:example}
\end{figure*}

\begin{figure*}[t]
    \centering
    \includegraphics[width=0.98\linewidth]{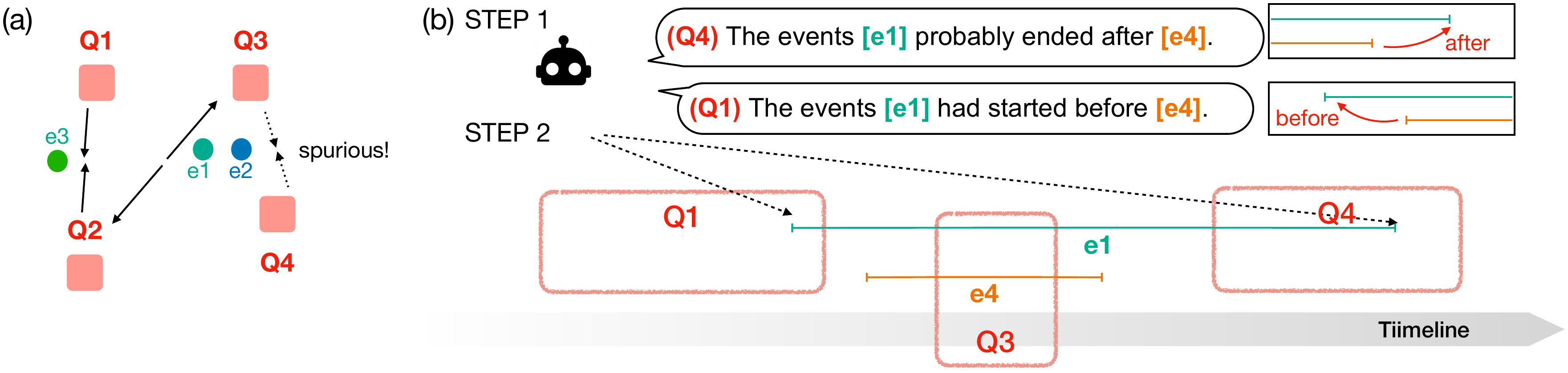}
    \vspace{-2mm}
    \caption{The illustration of (a) point-wise timeline grounding and (b) span-based one. (a) The model brings similar relations closer and pushes dissimilar ones apart, overlooking spurious overlap. (b) The speech bubbles in Step 1 describe the temporal evidence from each question-answer pair. The arrows in Step 2 describe the relative span prediction. It chains evidences about the timeline and mitigates spurious overlap.}
    \vspace{-5mm}
    \label{fig:timeline}
\end{figure*}

Understanding temporal relations is a challenging yet underexplored area in natural language processing
~\cite{ning-etal-2020-torque,chen2021dataset,zhou2019going}. \jonghoo{This challenge persists despite the prevalence of Large Language Models (LLMs)~\cite{chan2023chatgpt,fang2023getting}, whose training processes lack grounding in timeline evidence.}
One example task requiring such evidence is
temporal reading comprehension (TRC),
which requires to distinguish the temporal semantic difference between
\textit{``what \textbf{finished right before} the decision?''} and \textit{``what \textbf{finished right after} the decision?''.}

To distinguish the two questions, 
existing solution for TRC~\cite{Shang2021OpenTR} relies on
overlaps between related questions as a weak supervision to ground the semantics of temporal relations.
For example, in Figure \ref{fig:example}, if we let the question's target event as $X$, $Q1$ \textit{``what had \textbf{started before} $X$''} and $Q2$ \textit{``what \textbf{happened before} $X$''} have similar semantics ``\textit{\textbf{before}}''. Subsequently, the two share the overlapping answer \textit{``sent''}. On the other hand, the temporal semantics of $Q2$ and $Q3$ \textit{``what \textbf{happened while} $X$''} are different. So $Q2$ does not have any common answer with $Q3$.
By using answer overlaps as a proxy label, existing work proposes a contrastive objective.
It aims to pull the temporal relations in $Q1$ and $Q2$ closer together while broadening the distinction between $Q2$ and $Q3$.
This method performs comparably with or outperforms 
baselines requiring stronger but expensive human-annotations~\cite{han-etal-2021-econet, huang-etal-2022-understand}, as shown in Subsection~\ref{subsec:mainresults}.

\jonghoo{However, as illustrated in Figure~\ref{fig:timeline}, we argue that contrasting the evidence from answer overlaps misguide timeline as \textbf{point-wise} manner, leading to ``spurious overlap''.}
Questions $Q3$ and $Q4$ \textit{``What happened while X''} and \textit{``What probably ended after X''}, are temporally distinct but share answers ``taken'' and ``bearing''. 
In such cases, the point-wise timeline may 
fail to properly reason about the temporal meanings of the two questions. 
\jonghoo{The timeline mistakenly pulls $Q3$ and $Q4$ closer, making the model insufficient to differentiate the complex temporal questions.
The point-wise representation misses the timeline's inherent \textbf{span-based} nature.}

In this work, we focus on overcoming the limitations of point-wise event representation through span-based representations of time.
The key is utilizing the concept of time spans with notions of start and end points to supervise the complex temporal relationships between events.
For instance, the timeline in Figure~\ref{fig:timeline}(b) can separate $Q3$ and $Q4$
and distinguish between \textit{``happened while''} and \textit{``probably ended after''}, which are illustrated as disjoint boxes.
The overlap of the events is because the events span throughout the timeline, not because the questions are similar.
Despite its importance, previous work does not consider such a timeline due to the limited supervision in most scenarios.

. 
\jonghoo{We propose an advanced solution that elicits inductive reasoning behavior from a model grounded in predicted event spans.}
Inductive reasoning 
in the context of temporal relation understanding
is the process of extracting relations from individuals for deducing a whole, with the key purpose to acquire relative spans of events centered around a specific event.
First, the model answers each temporal relation
question in a ``question group'', a set of questions about the same event (e.g., $e4$).
As illustrated in speech bubbles in Figure \ref{fig:timeline}(b), \jonghoo{the question-answer pairs can be understood as part of the evidence about the timeline, such as `when event $e1$ occurred relative to the event $e4$'. 
Second, the model chains multiple temporal evidence within the same question group. 
This chained information forms a predicted timeline. For example, the speech bubbles in~Figure~\ref{fig:timeline}(b) collectively illustrate the start and end points of event $e1$.}
Supervised by the predicted timeline,
events that span a long time period can be identified, allowing us to discount attention to events with spurious overlaps.
 This process mitigates the spurious overlap without expensive human supervision.

Our model, Timeline Reasoning Network (\ours), equips the two-step inductive reasoning outlined as follows:
 An Evidence Extraction step aims to answer a specific question by extracting semantic and syntactic information with a pre-trained language model (PLM) and graph network.
 An Evidence Chaining step collectively predicts a timeline, using the novel attention module to chain multiple question-answer pairs. With the resulting timeline, the model grounds its answers consistently enhancing overall prediction accuracy.

We evaluate \ours on TORQUE and TB-Dense, a TRC and TRE task respectively.
We achieve state-of-the-art performance on the public leaderboard of TORQUE
~\footnote{\url{https://leaderboard.allenai.org/torque/submissions/public}}. We quantitatively and qualitatively analyze \ours's effectiveness in dealing with spurious overlaps,
which is measured by our new proposed ``passage level consistency'' metric.
Lastly, we confirm its generalizability on TB-Dense.
\jonghoo{
Our main contributions are three-fold:
\begin{itemize}
    \item We point out the spurious overlap issue in temporal relations, which arises from point-wise timeline grounding.
    \item We propose the inductive solution that chains evidence for the timeline in a span-based approach.
    \item Our novel framework, \ours, outperforms other approaches by effectively capturing temporal relations of events.
\end{itemize}
}

\section{Related Work}
We overview state-of-the-art works on temporal relation understanding and graph networks.


\paragraph{Temporal relation understanding}
Temporal relation understanding remains a challenging task even for large language models (LLMs)~\cite{chan2023chatgpt}. This includes task types such as TRE and TRC.
TRE tasks~\cite{Cassidy2014AnAF, Ning2018AMA} are to categorize the temporal order into pre-defined categories.
MATRES~\cite{Ning2018AMA} groups the temporal relations into  4 categories: \textit{Before/After/Simultaneous/Vague}. TB-Dense~\cite{Cassidy2014AnAF} considers 2 more classes, \textit{Includes} and \textit{Is Included}. Our proposed approach can benefit these tasks as we discuss in Section~\ref{sec:analysis}. 

Meanwhile, our main task is
the TRC task TORQUE~\cite{ning-etal-2020-torque}, requiring a temporal ordering in question form to reflect the real-world diversity of temporal relations.
Previous approaches to the TRC task include continuous pre-training~\cite{han-etal-2021-econet} and question decomposition methods~\cite{huang-etal-2022-understand, Shang2021OpenTR}. 
ECONET~\cite{han-etal-2021-econet} continually pre-trains the model to inject the knowledge of temporal orders.
Question decomposition approaches~\cite{huang-etal-2022-understand, Shang2021OpenTR} divide the question into the event part and temporal relation expression part to better capture the complex semantics. 
All of the above use contrastive methods to understand different temporal relations, either by contrasting relations with 
human annotations~\cite{han-etal-2021-econet,huang-etal-2022-understand} or annotated answers~\cite{Shang2021OpenTR}. 
However, the former can be costly or imprecise, while the latter may rely on spurious problems. 
Our distinction is the best of the two: no costly human annotation while avoiding spurious overlaps using span-based inductive reasoning.

\paragraph{Graph networks}

Graph Networks \cite{Kipf2016SemiSupervisedCW, Velickovic2017GraphAN} 
learn features through message passing on graph structures.
These networks have demonstrated their effectiveness in tasks requiring complex reasoning skills, such as numerical reasoning \cite{ran2019numnet, Chen2020QuestionDG} and logical reasoning \cite{huang2021dagn}.
Graph networks also have been applied to TRE \cite{Cheng2017ClassifyingTR, mathur-etal-2021-timers, zhang-etal-2022-extracting}, though their effectiveness in TRC has not been investigated.
\section{Proposed Method}
\begin{figure*}[t]
    \centering
    \includegraphics[width=0.90\linewidth]{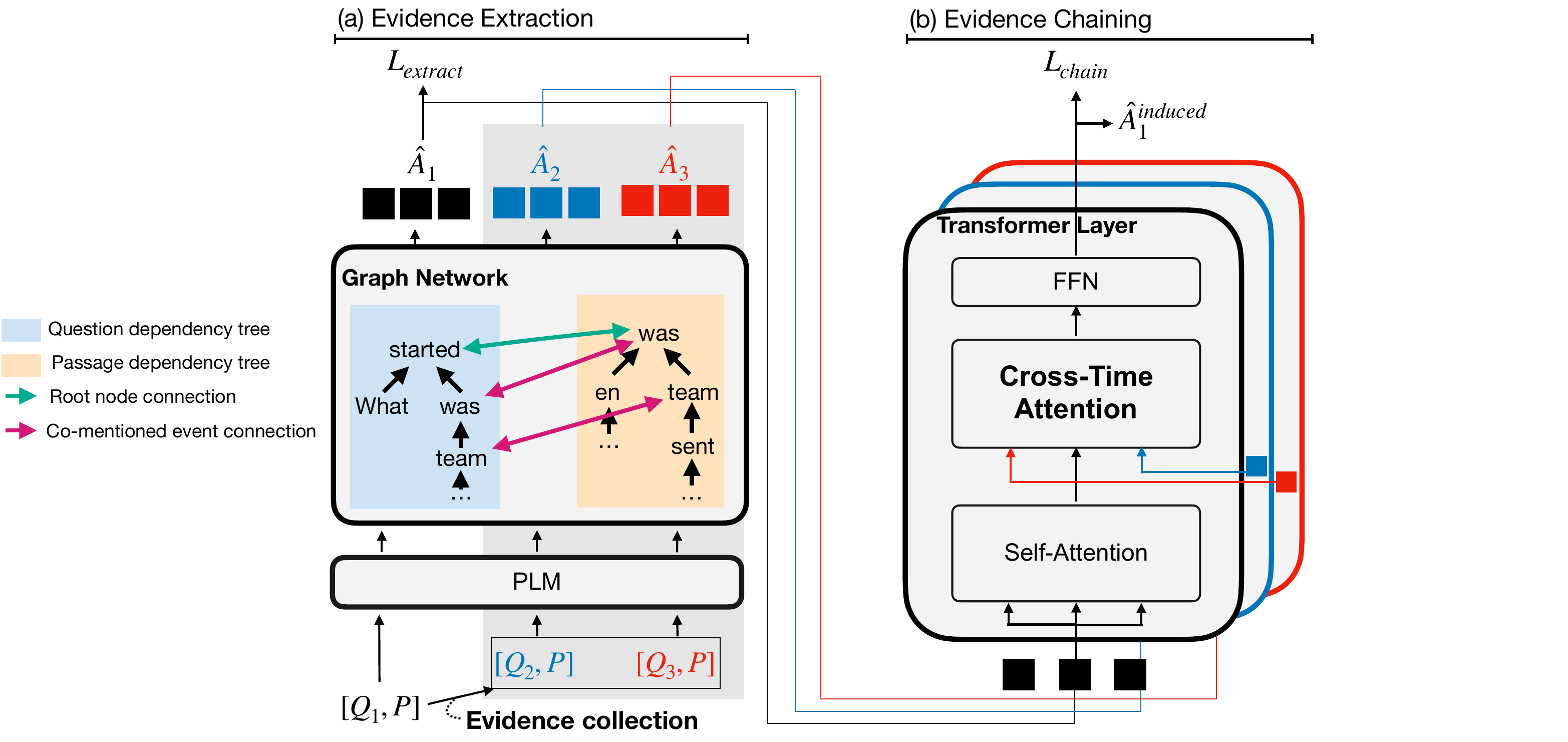}
    \vspace{-2mm}
    \caption{Overview of \ours. (a) The \graph answers each question with semantic (PLM) and syntactic (Graph Network) features. The example in the graph is from $Q1$ in Figure~\ref{fig:example}. (b) The \batchattn collects the related answers in the evidence collection stage and chains them through the cross-time attention module.}
    \vspace{-5mm}
    \label{fig:overall}
\end{figure*}

We formulate predicting answers for a query $Q$ as a binary classification for every word $p$ in the given passage $P$,
determining whether it is an answer event to $Q$~\footnote{
To facilitate a fair comparison with the available baselines in Section~\ref{sec:experimentalresults}, we also adopted the practice of using the first token as a word if a word is split into multiple tokens.
}.

Our approach is to solve the task with the two steps of inductive reasoning. The core of inductive reasoning is inferring the whole picture from individual evidence.
To transform the conventional function used in reading comprehension into the inductive form, we modify the function to consider answers to multiple questions together.
The conventional one is denoted as $\hat{A}_i = f(Q_i, P; \theta)$, where we answer ($\hat{A}_i$) the $i$-th question ($Q_i$) in the passage with model $\theta$. For inductive reasoning, the function is modified as:
\begin{equation}
\label{eq:inductiveform}
\begin{aligned}
\hat{A}^{induced}_i &= f(Q_i, P, \hat{A}^*; \theta), \\
\text{where } & \hat{A}^* = \{\hat{A}_i\}^l_{i=1}
\end{aligned}
\end{equation}
$l$ is the number of questions, and $\hat{A}^*$ is the set of model predictions for multiple questions.

The overview of our model is in Figure~\ref{fig:overall}. We first extract each answer ($\hat{A}_i$) as individual evidence in the Evidence Extraction step (Subsection~\ref{subsec:graph}), represented as the output squares in (a). The inductive reasoning is elicited in the Evidence Chaining step (Subsection~\ref{subsec:delib}).
We chain the related question-answers ($\hat{A}^*$) depicted as paths of blue and red, marked with a dark background, and utilize them in (b).

\subsection{\graph}
\label{subsec:graph} 

\jonghoo{The evidence extraction step aims to extract timeline evidence by answering each question.
We utilize both semantic information from PLM and syntactic information from the graph network.}
First, PLM encodes the question-passage pairs to get the contextual representation for each token.
It takes the concatenated sequence of pair as input $[Q, P]$ and outputs the vector representation $[Q^v$, $P^v]$, where each token is $q^v$ and $p^v$.

After that, we build a syntax-aware graph neural network that captures word-to-word dependency, which is an effective strategy for temporal reasoning~\cite{Cheng2017ClassifyingTR, mathur-etal-2021-timers, zhang-etal-2022-extracting}. 
Diverging from previous works mainly focused on temporal relations within passages and neglected questions, our formulation highlights the need to comprehend both.
As the graph in Figure~\ref{fig:overall}(a), we construct dependency tree graphs for both the question and passage, connecting root nodes and co-mentioned event words bidirectionally to facilitate the information exchange.
Here event words refer to nouns and verbs.

Next, we followed the graph reasoning step used in reading comprehension~\cite{ran2019numnet} that categorizes the connections of nodes into 4 types: (1) question-question $(qq)$ (2) passage-passage $(pp)$ (3) passage-question $(pq)$ (4) question-passage $(qp)$. Each node in the graph is the corresponding word in question and passage.
The pipeline consists of the following steps:

\begin{equation}
\label{eq:projection}
[\bar{Q},\bar{P}] = W^m [Q^v,P^v] + b^m 
\end{equation}
\begin{equation}
\label{eq:noderelated}
{\alpha_i = sigmoid(W^v \bar{v}_i + b^v)} 
\end{equation}
\begin{equation}
\label{eq:messageprop}
{\tilde{v}_i = \frac{1}{\left | N_i \right |}\left ( \sum_{}^{j \in N_i} \alpha_{j} W^{r_{ji}}\bar{v}[j] \right )}
\end{equation}
\begin{equation}
\label{eq:nodeupdate}
    v'_i = ReLU(W^u_i\bar{v}_i+{\tilde{v}}_i)+b^u
\end{equation}
(a) Projection: The vector outputs of the PLM pass through the projection layer $W^m$  
for node initialization (Eq.~\ref{eq:projection}).
(b) Node Relevance: We compute the weight $\alpha_i$ for each node $\bar{v}_i$ with the sigmoid function to determine the relevant nodes for answering temporal ordering questions (Eq.~\ref{eq:noderelated}). Here, nodes $\bar{v}$ consist of $\bar{q}$ and $\bar{p}$, each corresponding to the nodes from the question and passage.
(c) Message Propagation: The adjacency matrix $W^{r_{ji}}$ guides the message passing between nodes of different types (Eq.~\ref{eq:messageprop}), where $r_{ji} \in \{pp, pq, qp, qq\}$ and $N_i$ is the neighbor nodes of $\bar{v}_i$.
(d) Node Update: The message representations are added to the corresponding nodes, and a non-linear activation function (ReLU) is applied to update the node representations (Eq.~\ref{eq:nodeupdate}).

We iterate the steps (b), (c), and (d) for $T$ times.
Finally, the representation from PLM, $P^v$, is added and normalized to obtain the answer representations $\hat{A}_i$ in Eq.~\ref{eq:inductiveform}, with individual word representations $\hat{a}_i$.

\subsection{\batchattn}
\label{subsec:delib}
Our second and primary objective is to inductively reason with the group of questions and ground the answers with it. 
\jonghoo{A key motivation of the reasoning comes from the observation that chaining answers to questions about the same event serves as the relative timeline. Each prediction can be interpreted as temporal evidence like `when one event occurred relative to the asked event'. The pieces of evidence are then chained with the attention module to create the relative time span of passage events, helping the model ground its predictions.}

The evidence chaining step is built for such reasoning, whose process is further divided into two stages: evidence collection and timeline acquisition.
\paragraph{Evidence collection}
We first collect the question group, defined as questions that pertain to the same target event. Blue and red questions in Figure~\ref{fig:overall} correspond to the group. Task designs may provide the grouping for evaluation metrics~\cite{ning-etal-2020-torque} (Subsection~\ref{subsec:experimentalsetting}) or simple rules can be applied for the grouping (Subsection~\ref{subsec:tbdense}).

The questions are collectively encoded through the evidence extraction step and the output representations of them are collected.
If the model wants to answer the first question [$Q_1, P$] in the question group, the other questions $[Q_i, P]_{i=2}^{l}$ are encoded together to produce $[\hat{A}_i]_{i=2}^{l}$, then stacked with the original one to make $\{\hat{A}_i\}_{i=1}^{l}$, which corresponds to $\hat{A}^*$ in Eq.~\ref{eq:inductiveform}.

\paragraph{Timeline acquisition}
We need to build the timeline from the collected evidence to ensure the model's original answer is consistently grounded in such a timeline.

We achieve this through a novel transformer layer with our key component ``cross-time attention'' module.
Let the attention module $Attention(Q,K,V)$~\cite{vaswani2017attention}.
Conventional self-attention attends to tokens within a single data sequence-wise, represented by $Attention(p_{ki},p_{kj},p_{kj})$, where $k$ is the data index and $i,j$ are both equal or less than the sequence length. In contrast, our novel cross-time attention operates data-wise, gathering information from multiple data that were previously overlooked~\citep{vaswani2017attention}.
Each passage token $p_k$ attends to the same positioned token from related data. The equation of cross-time attention is:
\begin{equation}
    \scalebox{0.86}{$CrossTimeAttention = Attention(p_{ik},p_{jk},p_{jk})$}
\end{equation}
where $i,j \leq l$ and $k$ is token index.
We insert cross-time attention between the self-attention and feed-forward network (FFN) in the transformer layer.

In the evidence chaining step, the answer ($\hat{a}_i$) for the event ($p$) in $i$-th related question
conveys evidence of when event $p$ occurred relative to the event in question. Therefore, if the cross-time attention chains the pieces of temporal evidence of the event together, $\{\hat{a}_{i}\}_{i=1}^{l}$, it results in the time span of the event $p$.
The resulting time spans for events allow the model to refine the answer by collectively leveraging them as ground evidence. 

We enhance the model's reasoning behavior in temporal relation understanding through iterative application of our transformer layer $T'$ times.

\subsection{Training and Answer Prediction}
\label{subsec:losscalculation} 
At each step, the last output is fed to the one-layered perceptron head to get the prediction of whether the token is an answer to the question or not.
During the training phase, 
the final loss is the mean of extraction and chaining step losses, rewarding output from both steps.
The answer prediction loss from the first step, $L_{extract}$, guides the evidence from individual questions. The second step's loss, $L_{chain}$, guides the model to inductively correct the answer with the predicted timeline.
During the inference phase, our final logits, $\hat{A}^{induced}_i$, are the predictions of the evidence chaining step.
\newcommand\Tstrut{\rule{0pt}{2.2ex}}         
\newcommand\Bstrut{\rule[-0.9ex]{0pt}{0pt}}   

\begin{table*}[!ht]
\centering
\scalebox{0.88}{
\begin{tabular}{l|c|c|ccc|ccc}
\toprule
\multirow{2}{*}{Models}  & \multirow{2}{*}{wo external sup.} & \multirow{2}{*}{significance} & \multicolumn{3}{c|}{Dev}    & \multicolumn{3}{c}{Test} 
\Bstrut\\ 
\cline{4-9} 
& & & F1  & EM  & C & F1  & EM  & C  
\Tstrut\\ 

\midrule
RoBERTa-large & - & - & 75.7  & 50.4 & 36 & 75.2 & 51.1 & 34.5 \\ 
DeBERTa-large-v3 & - & - & 76.4  & 50.8 & 36.2 & 76.3 & 52.2 & 37.3 \\ 
ECONET & X & O & 76.9 & 52.2 & 37.7 & 76.3 & 52.0 & 37.0 \\ 
UBA  & X & X & 77.5  & 52.2  & 37.5  & 76.1  & 51.0  & \textbf{38.1}  \\ 
OTR-QA & O & X & 77.1 & 51.6 & \textbf{40.6} &  76.3 & 52.6 & 37.1  \\ 
\hline
 \ours (RoBERTa-large) &
  O & O & \textbf{77.6}$^{rd}$ &  
  \textbf{53.6}$^{rde}$ &
  40.3$^{rde}$ &
  \textbf{76.9}$^{rde}$ &
  \textbf{52.8}$^{rde}$ &
  \textbf{38.1}$^{r}$ \Tstrut\Bstrut \\ 
  \bottomrule
\end{tabular}
}
\vspace{-3mm}
\caption{Comparison between \ours and baselines on TORQUE dataset. 
We marked the models that (1) are trained without external supervision (2) have performed significance test on the test set.
Superscripts represent significant improvements compared to RoBERTa(r), DeBERTa(d) and ECONET(e). The best performance is denoted in bold.}

\vspace{-3mm}
\label{table:mainresult}
\end{table*}



\section{Experiment}
\label{sec:experimentalresults}

\subsection{Dataset and Evaluation Metrics}
\label{subsec:datasetandmetric}
We evaluate our proposed model on TORQUE dataset~\cite{ning-etal-2020-torque}, which is a temporal reading comprehension dataset.
It has 3.2k passages and 21.2k user-provided questions. 
Each instance has a question asking the temporal relationships between events described in a passage of text. 
TORQUE's annotation provides groups of questions, where one group consists of questions that were created by modifying the temporal nuance of an original seed question that dramatically changes the answers.
We use the official split~\footnote{\url{https://github.com/qiangning/TORQUE-dataset/tree/main}} and evaluation metrics, which include Macro F1, exact-match (EM), and consistency (C) as evaluation metrics.
C (consistency) is the percentage of question groups for which a model’s predictions have $ F1 \geq 80\% $ for all questions in a group.

\subsection{Baselines} 
 We compare our model against several baselines, including PLMs and models that use contrastive methods to teach the model temporal relations. Specifically,
\textbf{OTR-QA}~\cite{Shang2021OpenTR} reformulates the TORQUE task as open temporal relation extraction and uses answer overlap to weakly supervise temporal relations. As they target TORQUE without any external supervision like our method, they are our main baseline.
We further compare our model with those that use human-annotated temporal dictionaries.
\textbf{ECONET}~\cite{han-etal-2021-econet} is a continual pre-training approach with adversarial training that aims to equip models with knowledge about temporal relations. They use the external corpus and compile a dictionary of 40 common temporal expressions. 
\textbf{UBA}~\cite{huang-etal-2022-understand} employ the attention-based question decomposition to understand fine-grained questions. They also utilize a dictionary of temporal expressions as additional supervision, to capture the distinctions in temporal relationships.
\textbf{RoBERTa-large}~\cite{liu2019roberta} is a baseline PLM provided together with the TORQUE dataset and the previous SOTAs are based on. In addition, we evaluate the score of \textbf{DeBERTa-v3-large~\cite{he2022debertav3}}, which is known as the state-of-the-art PLM on a wide range of natural language understanding tasks.

\jonghoo{We don't regard recent LLMs as our main baseline due to their subpar performance in temporal relation understanding~\cite{chan2023chatgpt}. Additional evidence supporting this assessment is presented in our extended evaluation of ChatGPT in Appendix~\ref{sec:appendix:chatgpt}.}

\subsection{Experimental Settings}
\label{subsec:experimentalsetting}
 We search for optimized hyperparameters in our model. 
 $T$ and $T'$ are set between \{2, 3\} for the graph iteration step and for the evidence chaining step respectively.
 Each transformer layer in the evidence chaining step has 8 attention heads with a hidden size of 1024, and FFN layers in the attention module have dimensions between \{1024, 2048\}.
The question group is annotated for the C metric in TORQUE.
During the fine-tuning, 
the gradient accumulation step is set to 1, 
dropout ratio is set to 0.2 and other settings are identical with~\citet{ning-etal-2020-torque}. 
Spacy~\cite{spacy} is used for graph construction.
We use the PyTorch 1.11 library, and a NVIDIA GeForce RTX 3090 GPU with 42 average minutes to run an epoch.

For the performance report, we report the average score on the dev set and the best score on the test set to make a fair comparison with the baselines. This is because OTR-QA only reports the best single model results for all sets, and UBA reports single model results on the test set. 

For the significance test, we conducted paired t-tests ($p < 0.05$) only with PLMs and ECONET. It was due to the lack of reproducibility and significance test on the test set for OTR-QA and UBA. 

\subsection{Experimental Results}
\label{subsec:mainresults}
\autoref{table:mainresult} compares our approach to the baseline methods.
The baseline performances are provided by previous works \cite{ning-etal-2020-torque, han-etal-2021-econet, Shang2021OpenTR, huang-etal-2022-understand}.
The results show that \ours outperforms all compared baselines on both splits of TORQUE. 
\ours even surpasses ECONET and UBA which use a human-annotated dictionary of temporal expressions. 
Moreover, we found that while DeBERTa-v3-large shows a comparable score with OTR-QA, \ours significantly beats both DeBERTa-v3-large and OTR-QA. Such results indicate our approach shows notable benefits over existing methods.
One exception is the consistency score (C) of OTR-QA on the dev set.
But we note that \ours outperforms it in F1 and EM
and generalizes better to the test set, 
indicated by a much smaller dev-test gap in C (3.5 for OTR-QA vs 2.2 for \ours).
On the test set, \ours significantly outperforms all the baselines, achieving SOTA results on the TORQUE leaderboard.

\subsection{PLM variants}

\begin{table}[]
\centering
\scalebox{0.72}{
\begin{tabular}{l|ccc|ccc}
\toprule
\multirow{2}{*}{Models}   & \multicolumn{3}{c|}{Dev}    & \multicolumn{3}{c}{Test} 
\Bstrut\\ 
\cline{2-7} 
& F1  & EM  & C & F1  & EM  & C  
\Tstrut\\ 
\midrule
\multicolumn{7}{l}{DeBERTa-large-v3}                                                                                  \\ \hline
\multicolumn{1}{l|}{Naive}        & 76.4 & 50.8 & \multicolumn{1}{l|}{36.2} & 76.3          & 52.2          & 37.3 \\
\multicolumn{1}{l|}{TRN}          &         77.5 &    51.7  & \multicolumn{1}{l|}{39.1} & \textbf{77.3}          & \textbf{\textbf{53.5}} & \textbf{38.5} \\ \hline
\multicolumn{7}{l}{BERT-large}                                                                                     \\ \hline
\multicolumn{1}{l|}{Naive}        & 72.8 & 46.0   & \multicolumn{1}{l|}{30.7} & 71.9          & 45.9          & 29.1 \\
\multicolumn{1}{l|}{Current SOTA} & 73.5$\ddag$ & 46.5$\ddag$ & \multicolumn{1}{l|}{31.8$\ddag$} & \textbf{72.6$\ddag$}          & 45.1$\ddag$          & \textbf{30.1$\ddag$} \\
\multicolumn{1}{l|}{TRN}          & 73.1 & 47.2 & \multicolumn{1}{l|}{32.6} & 72.3          & \textbf{46.5}          & 29.8 \\ \hline
\multicolumn{7}{l}{RoBERTa-base}                                                                                   \\ \hline
\multicolumn{1}{l|}{Naive}        & 72.2 & 44.5 & \multicolumn{1}{l|}{28.7} & 72.6          & 45.7          & 29.9 \\
\multicolumn{1}{l|}{Current SOTA} &
  75.2$\dag$ &
  49.2$\dag$ &
  \multicolumn{1}{l|}{36.1$\dag$} &
  73.5$\ddag$ &
  \textbf{47.1$\dag$} &
  \textbf{32.7$\dag$} \\
\multicolumn{1}{l|}{TRN}          & 73.8 & 48.9 & \multicolumn{1}{l|}{34.7} & \textbf{73.7} & \textbf{47.1}          & 32.3 \\ \hline
\end{tabular}
}
\vspace{-3mm}
\caption{Comparison with PLM variants. Naive results of BERT-large and Roberta-base are from TORQUE~\cite{ning-etal-2020-torque} and DeBERTa-large from our own implementation. Current SOTA results are from OTR-QA~\cite{Shang2021OpenTR}$\dag$, UBA~\cite{huang-etal-2022-understand}$\ddag$.}
\label{table:plms}
\vspace{-3mm}
\end{table}

              
\autoref{table:plms} displays the results for PLM encoder variants. 
First, we implement our method on DeBERTa-v3-large~\cite{he2021debertav3} and observe that with the addition of \ours, it achieves the best test scores across all metrics. 
It demonstrates the effectiveness and generalizability of our method even with other PLM variants.
Our method is also shown to be generalizable to the BERT model, and its performance is comparable to other previous methods.
Lastly, when using the RoBERTa-base model, our results are again comparable to other baselines and surpass them in terms of F1 score, highlighting the scalability of \ours.
\begin{table}[]
\centering
\scalebox{0.83}{
\begin{tabular}{lccc}
\toprule
Models                     & \multicolumn{1}{l}{F1}   & \multicolumn{1}{l}{EM}   & \multicolumn{1}{l}{C} \\
\midrule
\ours  & \textbf{77.6} & \textbf{53.6}  & \textbf{40.3} \\ 
(d) \ours~- Self-Attention & 77.4 & 52.2  & 38.9 \\ 
(c) \ours~- Cross-Time Attention    & \multicolumn{1}{l}{76.3} & \multicolumn{1}{l}{51.4} & \multicolumn{1}{l}{38.6} \\ 
(b) \ours~- \batchattn & 76.0 & 51.9 & 38.1 \\ 
(a) \ours~- $G_{syn}$ & 76.1 & 50.9          & 37.3 \\ 
\bottomrule
\end{tabular}
}
\vspace{-2mm}
\caption{Ablation study on the dev set of TORQUE. Results are based on RoBERTa-large. The best performance is denoted in bold.}
\label{table:ablation}
\vspace{-5mm}
\end{table}

\subsection{Ablation Study}
\label{subsec:ablation}
To validate the effectiveness of each model component,
we conduct an ablation study on the dev set
and report the results 
in \autoref{table:ablation}.
In (a) we remove the syntactic graph network component $G_{syn}$ in the evidence extraction step and find the performance decreases significantly.
This suggests that syntactic graph reasoning helps
the downstream process of inductive reasoning
by creating passage token representations more effectively. 
For the evidence chaining step, 
we first remove 
(b) the whole layer, 
(c) the cross-time attention layer, 
and (d) the self-attention layer. 
The performance drops significantly with (b), 
indicating the importance of the evidence chaining step. 
Comparison between (c) and (d) indicates that 
the event chaining step helps performance gain by virtue of cross-time attention. 
It is the leading part of our reasoning elicitation by attending over the predicted timeline.
Meanwhile, (d) removing the simple stack of the transformer's self-attention part has the least impact on the performance.

\section{Discussion}
\label{sec:analysis}

While we empirically validated the effectiveness of \ours, its implication and generalizability can be
further clarified by the following discussion questions:
\begin{itemize}
    \item Q1: Does \ours mitigate spurious overlaps?
    \item Q2: Does \ours generalize to another task?
\end{itemize}


\subsection{Q1: Mitigating spurious overlaps}

As we have claimed comprehension of a span-based timeline works as a key constraint to avoid spurious overlaps,
we first address the question of whether the performance gain of \ours can be attributed to
a better comprehension of the timeline in the passage.

To quantitatively measure whether \ours understands passage timelines,
we adopt a passage-level consistency score $C_p$.
In TORQUE, each passage contains multiple question groups and each question group has questions asking about the same event. 
The original evaluation metric $C$ in~Subsection~\ref{subsec:datasetandmetric} measures consistency at a specific event or time point within the passage, by considering answer consistency in one question group. On the other hand, $C_p$ assesses the answer consistency across questions targeting different events by measuring the overall consistency 
of answers across multiple question groups within the same passage. We define $C_p$ as the percentage of passages for which a model’s predictions have $F1 \ge 80\%$ for all questions in a passage
~\footnote{We use the threshold of $F1 \geq 80\%$ for all the questions in the passage following the convention of \citet{gardner2020evaluating, ning-etal-2020-torque}.}. 

Through evaluating the consistency of answers across different time points corresponding to each target event, the $C_p$ score provides insights into the model's understanding of the time spans of events.
Therefore, if a model understands the passage timeline, its answers will be internally consistent with respect to the questions with different target events, which $C_p$ quantifies. We compare \ours with the model equipped with contrastive learning (\textit{CL}), which is implemented following OTR-QA's contrastive loss~\cite{Shang2021OpenTR}.

\begin{table}[]
\centering
\scalebox{0.85}{
\begin{tabular}{lcccc}
\toprule
Models      & \multicolumn{1}{l}{F1}   & \multicolumn{1}{l}{EM}   & \multicolumn{1}{l}{C}  & \multicolumn{1}{l}{$C_p$}   \\ 
\midrule
(a) Extract + Chain ~(\ours)  & \textbf{77.6}  & \textbf{53.6} & \textbf{40.3} &  \textbf{11.7} \\ 
(b) Chain  & 76.1  &   50.9  &  37.3  & 10.3 \\
(c) \textit{CL}  & 75.8 &  51.7 &  36.8  & 8.3 \\
\bottomrule
\end{tabular}
}
\vspace{-2mm}
\caption{Comparison of \textit{CL} and TDN on the dev set of TORQUE. The best performance is denoted in bold.}
\vspace{-5mm}
\label{table:versuscontrast}
\end{table}
 
\autoref{table:versuscontrast} shows that $C_p$ of \ours\ is significantly higher than that of \textit{CL}. 
To isolate the effect of the chaining step where the model reasons to predict the timeline, we also present ablated results removing the extraction step.
We observe that even
without the evidence extraction \ours outperforms \textit{CL}, which indicates that the improved understanding of
timeline plays a critical role in mitigating spurious overlap and thereby achieving performance gains
~\footnote{Though one may
argue adding the extraction step with \textit{CL} may further improve \textit{CL}, we found this was not the case (F1 and EM of 75.4 and 50.7 respectively), which is why we report CL itself.}.

\begin{figure}[h]
    \centering
    \scalebox{0.95}{
    \includegraphics[width=0.5\textwidth]{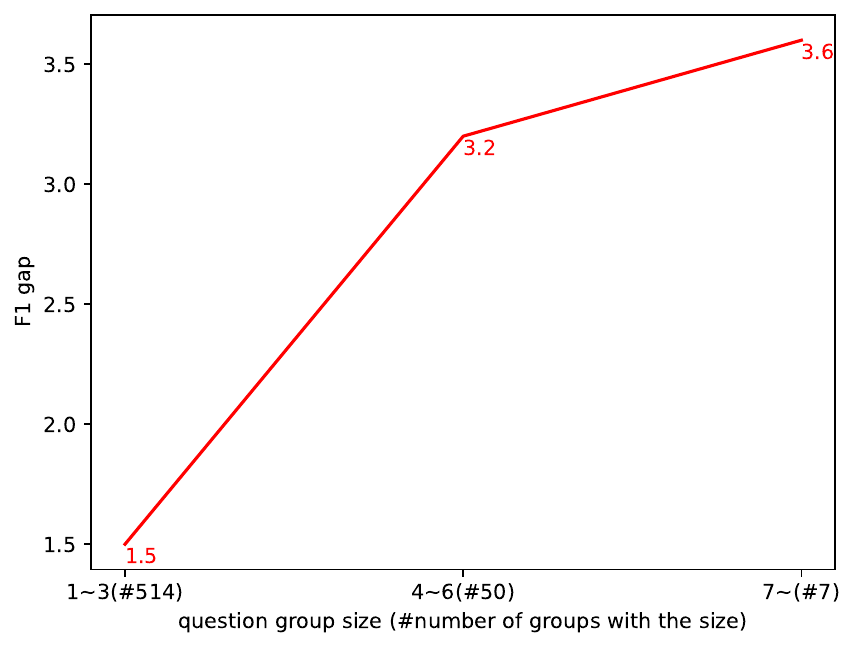}
    }
    \caption{Plot of the relationship between the question group size and F1 score gap. X-axis is the group size, binned into groups of 3. The number of groups in each bin is denoted in brackets. 
    Y-axis is the gap between the average F1 score of TRN and \textit{CL}, in percentage.}
    \vspace{-2mm}
    \label{fig:f1percluster}
\end{figure}

Figure~\ref{fig:f1percluster} groups F1 gains, by related question group sizes,
from which 
the gap from \textit{CL} widens as the size grows.
It is coherent with our hypothesis that \ours gains effectiveness by 
the timeline information predicted from multiple related questions,
which would be more effective for a larger question group size.
Moreover, our method persistently 
outperforms contrastive loss, even with a small question group size with a margin of $1.5pp$.



Lastly, as qualitative observations, Figure~\ref{fig:casestudy} in Appendix~\ref{sec:appendix:casestudy} compares answers from  \ours with \textit{CL}:
  \textit{CL} fails to clearly distinguish the semantic difference between Q1 and Q2, while our reasoning for the timeline avoids such mistakes.
  \ours is aware that
``exploded'' occurred before the tour ($Q3$), and not after the tour ($Q2$), so it 
cannot be during the same time as the tour ($Q4$). while
\textit{CL} fails. In addition, \ours finds the unmentioned events (e.g. ``arrested'' in $Q1$) and puts them in the right place on the timeline.

\begin{table}[]
\centering
\scalebox{0.85}{
\begin{tabular}{l|r|r}
\toprule
Models      & \multicolumn{1}{c|}{Dev} & \multicolumn{1}{c}{Test} \\  
\midrule
RoBERTa-large  & 58.9($\pm{2.1}$)                      & 63.4($\pm{2.3}$)                    \\
ECONET         & 60.8($\pm{0.6}$)                      & 64.8($\pm{1.4}$)                    \\ 
\midrule
\ours            & \textbf{62.5($\pm{1.4}$)}                   & \textbf{65.8($\pm{0.6}$)}            \\ 
\bottomrule
\end{tabular}
}
\caption{Micro-F1 scores on the TB-Dense dataset. The best performance on the test set is denoted in bold.}
\label{table:tbdense}
\vspace{-5mm}
\end{table}
\subsection{Q2: Generalization}
\label{subsec:tbdense}

To investigate whether our proposed approach generalizes to other temporal relation understanding task, we evaluate our method on TB-Dense~\cite{Cassidy2014AnAF}, which is a public benchmark for temporal relation extraction (TRE). 

For TB-Dense, when the passage and two event points in the passage are given,
the model must classify the relations between events into one of 6 types. 
As the explicit question is not provided in TB-Dense, we treat two event points as a question and group the questions in the dataset with a simple rule as follows: 
In the evidence extraction step, we prepend two events, $e1$, $e2$, to the passage $P$, and the model input is ``$[CLS] + e1 + e2 + [SEP] + P + [SEP]$''. 
In the evidence chaining step, we manually gather questions that are asked on the same first event within the same part of the passage, which can be easily identified by basic lexical matching. We use this gathering to construct the question group and predict the timeline.
We implement our method based on the publicly available source code of ECONET~\cite{han-etal-2021-econet}~\footnote{https://github.com/PlusLabNLP/ECONET}. 
Hyperparameters for fine-tuning are the same as ECONET.
The averages and standard deviations of Micro-F1 scores are reported from the runs with 3 different seeds. Since ECONET is the only model that targets both TORQUE and TB-Dense, we compare our results with it.

Our method achieves an F1 score of 65.8\% on this task,
compared to a RoBERTa-large baseline that achieves an F1 score of 63.4\%.
Moreover, our method outperforms ECONET, which requires an external corpus unlike ours.
These results demonstrate that
\ours's ability to build and utilize a predicted timeline is effective at various temporal relation understanding tasks,
and as such, our method has broader applicability beyond TRC.
\section{Conclusion}
\jonghoo{We introduce a novel approach for temporal relation understanding, which elicits inductive 
reasoning behaviors by predicting time spans of events.
Specifically, \ours~is collectively supervised from a span-based timeline built from multiple questions on the same event, as stronger evidence than answer overlaps that spuriously lead to point-wise timeline.}

\ours\ consists of the evidence extraction step that extracts individual evidence by answering each question with syntactic and semantic features, and the evidence chaining step that performs inductive reasoning for timeline prediction through a novel attention mechanism. 
Results on TORQUE and TB-dense datasets demonstrate that \ours outperforms previous methods by effectively mitigating the spurious answer overlaps. 

\section{Limitations}
Despite the promising results, there are some limitations to our approach. 
One limitation is that since we rely on the predicted timeline to mitigate spurious overlaps, it still has a chance of error. knowledge distillation or meta-learning could be applied in the future to remove the potential error.
Another limitation is that our main target, temporal reading comprehension, while a more realistic setting, is not commonly encountered in current NLP tasks. 
However, we argue that this is an important area that needs more active research,
especially considering applications of NLP models in real-world and real-time scenarios.
Moreover, while our primary focus has been on advancing methods for temporal understanding, it is important to highlight that our approach extends beyond this specific domain such as logical and causal reasoning. These domains share a common thread of requiring inductive reasoning skills, demonstrating the applicability of our proposed method.

For potential risks, our approach does not pose any significant risks. However, we note that
our work utilizes PLMs so biases may exist in the models due to the nature of their training data.

%


\section*{Acknowledgement}

This work was supported by Institute of Information \& communications Technology Planning \& Evaluation (IITP) grant funded by the Korea government(MSIT) [NO.2021-0-01343, Artificial Intelligence Graduate School Program (Seoul National University)] and  Institute of Information \& Communications Technology Planning \& Evaluation (IITP) grant funded by the Korean government (MSIT) (No. 2022-0-00077, AI Technology Development for Commonsense Extraction, Reasoning, and Inference from Heterogeneous Data).
\bibliography{anthology,custom}

\begin{thebibliography}{24}
\expandafter\ifx\csname natexlab\endcsname\relax\def\natexlab#1{#1}\fi

\bibitem[{Cassidy et~al.(2014)Cassidy, McDowell, Chambers, and
  Bethard}]{Cassidy2014AnAF}
Taylor Cassidy, Bill McDowell, Nathanael Chambers, and Steven Bethard. 2014.
\newblock An annotation framework for dense event ordering.
\newblock In \emph{Annual Meeting of the Association for Computational
  Linguistics}.

\bibitem[{Chan et~al.(2023)Chan, Cheng, Wang, Jiang, Fang, Liu, and
  Song}]{chan2023chatgpt}
Chunkit Chan, Jiayang Cheng, Weiqi Wang, Yuxin Jiang, Tianqing Fang, Xin Liu,
  and Yangqiu Song. 2023.
\newblock \href {http://arxiv.org/abs/2304.14827} {Chatgpt evaluation on
  sentence level relations: A focus on temporal, causal, and discourse
  relations}.

\bibitem[{Chen et~al.(2020)Chen, Xu, Cheng, Xiaochuan, Zhang, Song, Wang, Qi,
  and Chu}]{Chen2020QuestionDG}
Kunlong Chen, Weidi Xu, Xingyi Cheng, Zou Xiaochuan, Yuyu Zhang, Le~Song,
  Taifeng Wang, Yuan Qi, and Wei Chu. 2020.
\newblock Question directed graph attention network for numerical reasoning
  over text.
\newblock In \emph{Conference on Empirical Methods in Natural Language
  Processing}.

\bibitem[{Chen et~al.(2021)Chen, Wang, and Wang}]{chen2021dataset}
Wenhu Chen, Xinyi Wang, and William~Yang Wang. 2021.
\newblock A dataset for answering time-sensitive questions.
\newblock \emph{arXiv preprint arXiv:2108.06314}.

\bibitem[{Cheng and Miyao(2017)}]{Cheng2017ClassifyingTR}
Fei Cheng and Yusuke Miyao. 2017.
\newblock Classifying temporal relations by bidirectional lstm over dependency
  paths.
\newblock In \emph{Annual Meeting of the Association for Computational
  Linguistics}.

\bibitem[{Fang et~al.(2023)Fang, Wang, Zhou, Zhang, Song, and
  Chen}]{fang2023getting}
Tianqing Fang, Zhaowei Wang, Wenxuan Zhou, Hongming Zhang, Yangqiu Song, and
  Muhao Chen. 2023.
\newblock \href {http://arxiv.org/abs/2305.14970} {Getting sick after seeing a
  doctor? diagnosing and mitigating knowledge conflicts in event temporal
  reasoning}.

\bibitem[{Gardner et~al.(2020)Gardner, Artzi, Basmov, Berant, Bogin, Chen,
  Dasigi, Dua, Elazar, Gottumukkala et~al.}]{gardner2020evaluating}
Matt Gardner, Yoav Artzi, Victoria Basmov, Jonathan Berant, Ben Bogin, Sihao
  Chen, Pradeep Dasigi, Dheeru Dua, Yanai Elazar, Ananth Gottumukkala, et~al.
  2020.
\newblock Evaluating models’ local decision boundaries via contrast sets.
\newblock \emph{Findings of Empirical Methods in Natural Language Processing}.

\bibitem[{Han et~al.(2021)Han, Ren, and Peng}]{han-etal-2021-econet}
Rujun Han, Xiang Ren, and Nanyun Peng. 2021.
\newblock \href {https://doi.org/10.18653/v1/2021.emnlp-main.436} {{ECONET}:
  Effective continual pretraining of language models for event temporal
  reasoning}.
\newblock In \emph{Proceedings of the 2021 Conference on Empirical Methods in
  Natural Language Processing}, pages 5367--5380, Online and Punta Cana,
  Dominican Republic. Association for Computational Linguistics.

\bibitem[{He et~al.(2021)He, Gao, and Chen}]{he2021debertav3}
Pengcheng He, Jianfeng Gao, and Weizhu Chen. 2021.
\newblock Debertav3: Improving deberta using electra-style pre-training with
  gradient-disentangled embedding sharing.
\newblock \emph{arXiv preprint arXiv:2111.09543}.

\bibitem[{He et~al.(2022)He, Gao, and Chen}]{he2022debertav3}
Pengcheng He, Jianfeng Gao, and Weizhu Chen. 2022.
\newblock Debertav3: Improving deberta using electra-style pre-training with
  gradient-disentangled embedding sharing.
\newblock In \emph{The Eleventh International Conference on Learning
  Representations}.

\bibitem[{Honnibal et~al.(2020)Honnibal, Montani, Van~Landeghem, and
  Boyd}]{spacy}
Matthew Honnibal, Ines Montani, Sofie Van~Landeghem, and Adriane Boyd. 2020.
\newblock \href {https://doi.org/10.5281/zenodo.1212303} {spacy:
  Industrial-strength natural language processing in python}.

\bibitem[{Huang et~al.(2022)Huang, Geng, Long, and
  Jiang}]{huang-etal-2022-understand}
Hao Huang, Xiubo Geng, Guodong Long, and Daxin Jiang. 2022.
\newblock \href {https://doi.org/10.18653/v1/2022.naacl-main.28} {Understand
  before answer: Improve temporal reading comprehension via precise question
  understanding}.
\newblock In \emph{Proceedings of the 2022 Conference of the North American
  Chapter of the Association for Computational Linguistics: Human Language
  Technologies}, pages 375--384, Seattle, United States. Association for
  Computational Linguistics.

\bibitem[{Huang et~al.(2021)Huang, Fang, Cao, Wang, and Liang}]{huang2021dagn}
Yinya Huang, Meng Fang, Yu~Cao, Liwei Wang, and Xiaodan Liang. 2021.
\newblock Dagn: Discourse-aware graph network for logical reasoning.
\newblock In \emph{Proceedings of the 2021 Conference of the North American
  Chapter of the Association for Computational Linguistics: Human Language
  Technologies}, pages 5848--5855.

\bibitem[{Kipf and Welling(2016)}]{Kipf2016SemiSupervisedCW}
Thomas Kipf and Max Welling. 2016.
\newblock Semi-supervised classification with graph convolutional networks.
\newblock \emph{ArXiv}, abs/1609.02907.

\bibitem[{Liu et~al.(2019)Liu, Ott, Goyal, Du, Joshi, Chen, Levy, Lewis,
  Zettlemoyer, and Stoyanov}]{liu2019roberta}
Yinhan Liu, Myle Ott, Naman Goyal, Jingfei Du, Mandar Joshi, Danqi Chen, Omer
  Levy, Mike Lewis, Luke Zettlemoyer, and Veselin Stoyanov. 2019.
\newblock Roberta: A robustly optimized bert pretraining approach.
\newblock \emph{arXiv preprint arXiv:1907.11692}.

\bibitem[{Mathur et~al.(2021)Mathur, Jain, Dernoncourt, Morariu, Tran, and
  Manocha}]{mathur-etal-2021-timers}
Puneet Mathur, Rajiv Jain, Franck Dernoncourt, Vlad Morariu, Quan~Hung Tran,
  and Dinesh Manocha. 2021.
\newblock \href {https://doi.org/10.18653/v1/2021.acl-short.67} {{TIMERS}:
  Document-level temporal relation extraction}.
\newblock In \emph{Proceedings of the 59th Annual Meeting of the Association
  for Computational Linguistics and the 11th International Joint Conference on
  Natural Language Processing (Volume 2: Short Papers)}, pages 524--533,
  Online. Association for Computational Linguistics.

\bibitem[{Ning et~al.(2020)Ning, Wu, Han, Peng, Gardner, and
  Roth}]{ning-etal-2020-torque}
Qiang Ning, Hao Wu, Rujun Han, Nanyun Peng, Matt Gardner, and Dan Roth. 2020.
\newblock \href {https://doi.org/10.18653/v1/2020.emnlp-main.88} {{TORQUE}: A
  reading comprehension dataset of temporal ordering questions}.
\newblock In \emph{Proceedings of the 2020 Conference on Empirical Methods in
  Natural Language Processing (EMNLP)}, pages 1158--1172, Online. Association
  for Computational Linguistics.

\bibitem[{Ning et~al.(2018)Ning, Wu, and Roth}]{Ning2018AMA}
Qiang Ning, Hao Wu, and Dan Roth. 2018.
\newblock A multi-axis annotation scheme for event temporal relations.
\newblock In \emph{Annual Meeting of the Association for Computational
  Linguistics}.

\bibitem[{Ran et~al.(2019)Ran, Lin, Li, Zhou, and Liu}]{ran2019numnet}
Qiu Ran, Yankai Lin, Peng Li, Jie Zhou, and Zhiyuan Liu. 2019.
\newblock {N}um{N}et: Machine reading comprehension with numerical reasoning.
\newblock In \emph{Proceedings of the 2019 Conference on Empirical Methods in
  Natural Language Processing and the 9th International Joint Conference on
  Natural Language Processing (EMNLP-IJCNLP)}, pages 2474--2484.

\bibitem[{Shang et~al.(2021)Shang, Qi, Wang, Huang, Wu, and
  Zhou}]{Shang2021OpenTR}
Chao Shang, Peng Qi, Guangtao Wang, Jing Huang, Youzheng Wu, and Bowen Zhou.
  2021.
\newblock Open temporal relation extraction for question answering.
\newblock In \emph{Conference on Automated Knowledge Base Construction}.

\bibitem[{Vaswani et~al.(2017)Vaswani, Shazeer, Parmar, Uszkoreit, Jones,
  Gomez, Kaiser, and Polosukhin}]{vaswani2017attention}
Ashish Vaswani, Noam Shazeer, Niki Parmar, Jakob Uszkoreit, Llion Jones,
  Aidan~N Gomez, {\L}ukasz Kaiser, and Illia Polosukhin. 2017.
\newblock Attention is all you need.
\newblock \emph{Advances in neural information processing systems}, 30.

\bibitem[{Velickovic et~al.(2017)Velickovic, Cucurull, Casanova, Romero,
  Lio’, and Bengio}]{Velickovic2017GraphAN}
Petar Velickovic, Guillem Cucurull, Arantxa Casanova, Adriana Romero, Pietro
  Lio’, and Yoshua Bengio. 2017.
\newblock Graph attention networks.
\newblock \emph{ArXiv}, abs/1710.10903.

\bibitem[{Zhang et~al.(2022)Zhang, Ning, and
  Huang}]{zhang-etal-2022-extracting}
Shuaicheng Zhang, Qiang Ning, and Lifu Huang. 2022.
\newblock \href {https://doi.org/10.18653/v1/2022.findings-naacl.29}
  {Extracting temporal event relation with syntax-guided graph transformer}.
\newblock In \emph{Findings of the Association for Computational Linguistics:
  NAACL 2022}, pages 379--390, Seattle, United States. Association for
  Computational Linguistics.

\bibitem[{Zhou et~al.(2019)Zhou, Khashabi, Ning, and Roth}]{zhou2019going}
Ben Zhou, Daniel Khashabi, Qiang Ning, and Dan Roth. 2019.
\newblock “going on a vacation” takes longer than “going for a walk”: A
  study of temporal commonsense understanding.
\newblock In \emph{Proceedings of the 2019 Conference on Empirical Methods in
  Natural Language Processing and the 9th International Joint Conference on
  Natural Language Processing (EMNLP-IJCNLP)}, pages 3363--3369.

\end{thebibliography}

\appendix


\section{ChatGPT evaluation}
\label{sec:appendix:chatgpt}

\begin{table*}[!ht]
\centering
\begin{tabular}{l|ccc|ccc}
\toprule
\multirow{2}{*}{Models}   & \multicolumn{3}{c|}{Dev}    & \multicolumn{3}{c}{Test} 
\Bstrut\\ 
\cline{2-7} 
& F1  & EM  & C & F1  & EM  & C  
\Tstrut\\ 
\midrule
\multicolumn{1}{l|}{Fine-tuned RoBERTa-large}        & 75.7 & 50.4 & \multicolumn{1}{l|}{36} & 75.2          & 51.1         & 34.5 \\
\midrule
\multicolumn{1}{l|}{ChatGPT ICT$_{1 shot}$}        & 28.2 & 26.1 & \multicolumn{1}{l|}{2.7} & 30.4          & 28.4          & 3.5 \\
\multicolumn{1}{l|}{ChatGPT ICT$_{3 shot}$}          &    33.5  &   29.4  & \multicolumn{1}{l|}{3.5} & 34.3          &  30.9 & 3.5 \\ \hline
\end{tabular}
\caption{The performance of ChatGPT vs fine-tuned RoBERTa on TORQUE dataset. }
\label{table:chatgpteval}
\end{table*}
\jonghoo{
Table~\ref{table:chatgpteval} presents the evaluation results of ChatGPT (gpt-3.5-turbo-0301) on the TORQUE dataset. The outcomes reveal that the performance of ChatGPT on the TRC task is significantly inferior to that of a fine-tuned model, consistent with the observation that ChatGPT performs poorly on TRE tasks~\cite{chan2023chatgpt}.

Regarding the evaluation settings, we introduced prompts for TORQUE as elaborated in Table~\ref{table:chatgptprompt} and conducted in-context learning (ICT). Then the model outputs are separated by commas to obtain the answers.
}

\newcommand{\specialcell}[2][c]{%
  \begin{tabular}[#1]{@{}c@{}}#2\end{tabular}}

\begin{table*}
\small
\centering
\begin{tabular}{c|c}
\toprule
\multicolumn{1}{c|}{Strategies} & Template input \\
\midrule
\specialcell{1-shot}  & \parbox[c]{9cm}{\texttt{Question: }What had started before a woman was trapped? \texttt{Select answer events from the passage. One event corresponds to exactly one word. If there are no events, select None.} \\ \texttt{Passage: }Heavy snow is causing disruption to transport across the UK, with heavy rainfall bringing flooding to the south-west of England. Rescuers searching for a woman trapped in a landslide at her home in Looe, Cornwall, said they had found a body.\\ \texttt{Answer: }snow, rainfall, landslide\\ \\
\texttt{Question: }What happened before something was not mentioned? 
\texttt{Select answer events from the passage. One event corresponds to exactly one word. If there are no events, select None.}\\ 
\texttt{Passage: }Titled ``Beyond Human'', the script ``threw in a lot about UFOs and space aliens and earthlings evolving from their `containers' to a `higher level,''' Papas said. Suicide is not mentioned in the script, he added.\\
\texttt{Answer:}
} \\
\midrule
\specialcell{3-shot}  & \parbox[c]{9cm}{
\texttt{Question: }What had started before a woman was trapped? 
\texttt{Select answer events from the passage. One event corresponds to exactly one word. If there are no events, select None.}\\ 
\texttt{Passage: }Heavy snow is causing disruption to transport across the UK, with heavy rainfall bringing flooding to the south-west of England. Rescuers searching for a woman trapped in a landslide at her home in Looe, Cornwall, said they had found a body.\\ 
\texttt{Answer: }snow, rainfall, landslide\\ \\ 
\texttt{Question: }What happened before Lisa Schlein reports? 
\texttt{Select answer events from the passage. One event corresponds to exactly one word. If there are no events, select None.}\\ 
\texttt{Passage: }Special events are being organized by the European Commission and individual nations in Europe, North America, and other parts of the world. Lisa Schlein reports from Geneva.\\ 
\texttt{Answer: }None\\ \\ 
\texttt{Question: }What could have happened after the votes were provided? 
\texttt{Select answer events from the passage. One event corresponds to exactly one word. If there are no events, select None.}\\ 
\texttt{Passage: }But instead of providing the votes to strike it down, they chose to uphold it on the flimsy ground that because the sex of the parent and not the child made the difference under the law, the plaintiff did not have standing to bring the case. The Justice Department, which supported the statute, did not cover itself with glory either.\\ 
\texttt{Answer: }strike, have, bring, cover \\ \\
\texttt{Question: }What happened before something was not mentioned? 
\texttt{Select answer events from the passage. One event corresponds to exactly one word. If there are no events, select None.}\\ 
\texttt{Passage: }Titled ``Beyond Human'', the script ``threw in a lot about UFOs and space aliens and earthlings evolving from their `containers' to a `higher level,''' Papas said. Suicide is not mentioned in the script, he added.\\
\texttt{Answer:}
} \\
\bottomrule
\end{tabular}
\caption{ChatGPT prompt templates used for the TORQUE dataset.}
\label{table:chatgptprompt}
\end{table*}
\section{Qualitative analysis}
\label{sec:appendix:casestudy}

\begin{figure*}[h]
    \centering
    \includegraphics[width=1\textwidth]{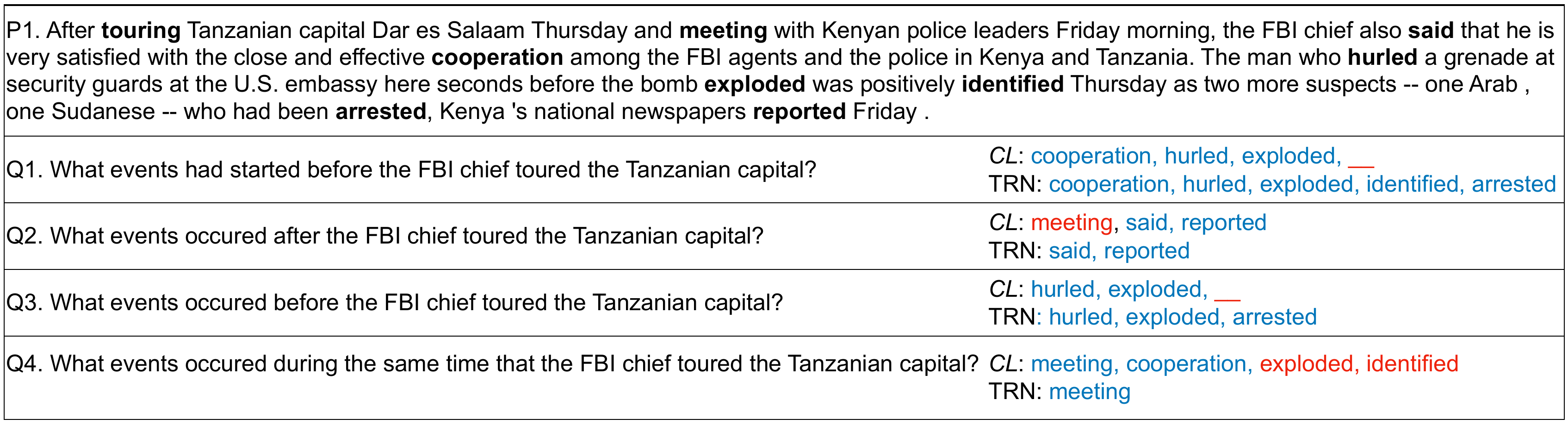}
    \vspace{-7mm}
    \caption{Qualitative analysis of contrastive learning and \ours. Events in the passage are highlighted in bold. In answers, correct events are denoted in blue, and incorrect events are denoted in red. Missing events are underlined.}
    \vspace{-5mm}
    \label{fig:casestudy}
\end{figure*}

Figure~\ref{fig:casestudy} shows the qualitative analysis that compares the predictions of \ours with contrastive learning.

\end{document}